\makeatletter
\providecommand{\@trackname}{}
\makeatother
\documentclass{article}

\usepackage[final]{neurips_2026}
\usepackage[utf8]{inputenc}
\usepackage[T1]{fontenc}
\usepackage{hyperref}
\usepackage{url}
\usepackage{booktabs}
\usepackage{amsmath}
\usepackage{graphicx}
\usepackage{flafter}
\usepackage{placeins}
\usepackage{xcolor}
\usepackage{CJKutf8}

\setcitestyle{authoryear,round}

\title{Cross-Lingual Persona Effects in LLM Negotiation}
\workshoptitle{SocialAgent: Second Workshop on Large Language Models for Social Reasoning and Simulation}
\author{%
  Saehee Eom \\
  Georgia Institute of Technology \\
  \texttt{seom@gatech.edu} \\
  \And
  Tejas Khandwekar \\
  Georgia Institute of Technology \\
  \texttt{tkhandwekar3@gatech.edu} \\
}

\begin{document}
\maketitle

\begin{abstract}
Tell an LLM negotiator to be ``headstrong'' and it should hold firm on
price, but does that instruction retain the same behavior in every language? To test this, we use identical buyer and seller instances of Gemini 3.6 Flash
to negotiate the same 12 CraigslistBargain scenarios in English and Korean.
The pilot-scale $2\times2\times3$ design crosses interaction language, seller
urgency (none or strong), and buyer persona (cooperative, neutral, or
headstrong), yielding 432 attempted negotiations. We call this property \emph{cross-lingual
social-policy fidelity}: whether a prompt-defined social policy produces the
same behavioral effect regardless of interaction language. It often does
not. Korean negotiations reached agreement far more often than English
(77.3\% vs.\ 52.5\%, a 24.8-point gap, Holm-adjusted $p=.0039$), but only for
neutral and headstrong buyers; cooperative buyers were already near-ceiling
in both languages. Buyers' opening offers barely differed by language; the
gap instead emerged later, through larger seller concessions and a smaller
final gap in Korean, and agreed prices were also lower in Korean. Repeating
the full design with Qwen3.7 Flash as both agents gives directional support:
agreement was again higher in Korean ($+13.9$ points vs.\ Gemini's $+24.8$,
with an interval whose lower bound is near zero), and sellers again conceded
more. 
Pragmatic analysis reveals why: while both languages frequently featured face-threatening acts like rejections and counter-offers, Korean dialogue softened these threats with substantially more politeness mitigation, especially among sellers (11.7\% vs. 3.6\%).
These results show that such a policy can produce
different simulated negotiation behavior, not just different wording, when only the
interaction language changes.
\end{abstract}

\section{Introduction}
Negotiation is a compact test of socially situated agents: models must track
goals, read counterparts, choose offers, and decide whether to concede,
accept, or walk away. Early neural negotiators learned strategy and language
jointly \citep{lewis2017deal}; CraigslistBargain later decoupled them through
grounded buyer--seller dialogue acts \citep{he2018decoupling}. Recent LLM work
shows that tactics, roles, personality, and social framing also affect
negotiation outcomes \citep{bianchi2024negotiationarena,kwon2024negotiators,
xia2024bargaining,huang2024personality,noh2026personalities,vaccaro2026advancing}.

Language itself shapes negotiation outcomes: multilingual studies find that
agreement rates, stability, and reasoning behavior all shift with the
language used \citep{sinha2026language,hakimov2026price}, and prompt
politeness and social reasoning vary across languages, Korean included
\citep{yin2024respect,jung2024koreanpoliteness,kim2026socialreasoning}.
Yet language and social configuration are almost always studied separately,
leaving an open question for deployment: does a persona, or a response to a
counterpart's hardship, still work the same way once the interaction
language changes?

We answer this question with matched English--Korean self-play: the same 12
items, private targets, bargaining rules, model, and
decoding settings, crossed only with seller urgency and buyer persona.
Because every turn carries both an utterance and a structured action, we do
not have to infer strategy from the text. We can see exactly where anchors,
concessions, and the final buyer--seller gap actually move. We call this
target property \textbf{cross-lingual social-policy fidelity}: whether a
persona- or hardship-based policy keeps its behavioral effect across interaction
languages.

We evaluate cross-lingual social-policy fidelity at two levels. Structured
actions reveal where agreement, concessions, and terminal gaps diverge, while
a separate judge model characterizes face-threatening acts (FTAs), mitigation,
and politeness strategies across languages. These pragmatic annotations are secondary, descriptive evidence and do not establish a causal mechanism.

\section{Method}
\label{sec:method}

\subsection{Design}

We sampled 12 CraigslistBargain scenarios
\citep{he2018decoupling,stanfordnlp2026craigslist}, two from each of six item
categories. Each has a listing price and private buyer and seller targets.
Korean titles and descriptions were machine translated (Gemini 3.6 Flash) and checked line-by-line by one fluent Korean speaker on the author team. U.S. locations and dollar prices were kept, so the Korean condition is a
translated U.S. marketplace, not a Korean one (Appendix~\ref{app:translation}).
We crossed language (English, Korean), seller urgency (none, strong need for
cash), and buyer persona (cooperative, neutral, headstrong). Cooperative
buyers are told to concede readily and headstrong buyers to anchor on their
target and concede slowly; neutral adds nothing. Persona and urgency text is
written in the interaction language, so instruction and conversation language
change together, and buyer and seller always share a language. Each of the 12
conditions ran three times per scenario, for 432 negotiations.

Buyer and seller were separate \texttt{gemini-3.6-flash} instances at
temperature 0.2. The seller opens at the listing price and the roles alternate
for up to 16 turns. Each turn returns an utterance, an action (offer,
counter-offer, accept, reject), and an optional price; an accept ends the
negotiation in a deal and a reject ends it without one. Two English runs that
could not be parsed were dropped, leaving 430 (Appendix~\ref{app:protocol}).

\subsection{Measures}

The main outcome is agreement; we also report abandonment, turn-capped outcomes,
length, and agreed price, normalized as
$z(p)=(p-\text{buyer target})/(\text{seller target}-\text{buyer target})$ so
that 0 is the buyer's target and 1 the seller's. Agreed prices exist only for
deals and are therefore conditional on agreement. Process measures (opening
offers, net concessions by role, and the final buyer--seller gap) use every
run. A separate model, \texttt{qwen/qwen3.7-flash}, labeled each turn for
face-threatening acts (FTAs) and Brown--Levinson politeness strategy
\citep{brown1987politeness,dutt2020face}; we use these labels descriptively.

\subsection{Statistical analysis}


To weight each scenario equally, we average repetitions within each factorial
cell before averaging across scenarios. Confidence intervals are computed
using 10,000 scenario-block bootstrap samples. Four agreement contrasts were
prespecified and tested with exact two-sided sign-flip tests under Holm
correction: language, language$\times$urgency,
language$\times$(cooperative$-$neutral), and
language$\times$(headstrong$-$neutral). All other analyses are exploratory.

\section{Results}
\label{sec:results}

\subsection{Agreement outcomes}

Scenario-balanced agreement was 52.5\% in English (112/214 valid runs) and
77.3\% in Korean (167/216), a $+24.8$-point contrast (Table~\ref{tab:headline};
full per-cell counts are in Appendix~\ref{app:cells}, and per-scenario
counts in Appendix~\ref{app:scenario-agreement}).
Strong seller urgency also raised agreement from 50.0\% to 79.9\%
($+29.9$ points). The $+16.2$-point language$\times$urgency contrast has a
bootstrap interval above zero, but its exact test does not survive the
four-test Holm family ($p=.1133$).

\begin{figure}[t]
  \centering
  \includegraphics[width=.85\linewidth]{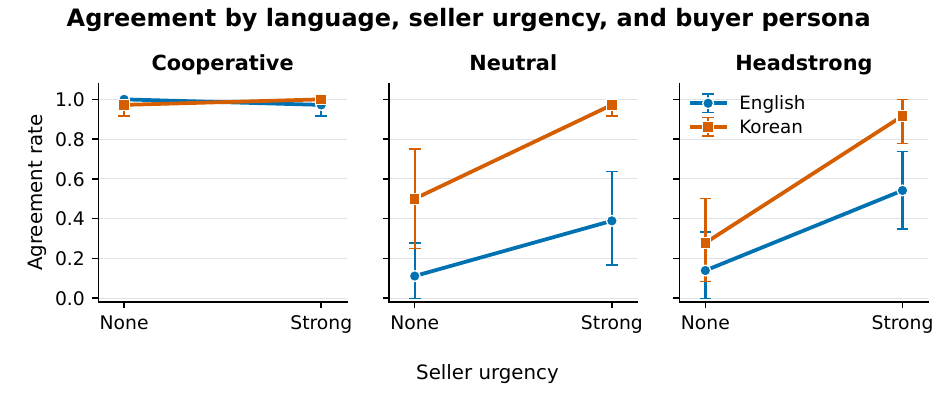}
  \caption{Scenario-balanced agreement by language, seller urgency, and buyer
  persona; bars are 95\% scenario-block bootstrap intervals. Cooperative
  agreement is near ceiling in both languages, while Korean most strongly
  changes neutral and headstrong outcomes.}
  \label{fig:agreement}
\end{figure}


Decomposing this overall gain reveals that it is concentrated in specific buyer personas rather than uniform across conditions (Figure~\ref{fig:agreement}). Cooperative buyers reached 98.6\% agreement in both languages, leaving little room for a language effect. Instead, the gap is driven by neutral buyers (rising from 25.0\% in English to 73.6\% in Korean) and headstrong buyers (rising from 34.0\% to 59.7\%). In particular, the English neutral buyer, no-urgency seller baseline was prone to negotiation breakdown, showing only 11.1\% agreement and 88.9\% abandonment. Accordingly, both language-by-persona contrasts remained statistically significant after Holm correction. Exploratory three-way interaction intervals crossed zero, indicating that seller urgency does not systematically alter these language-by-persona dynamics further.

\begin{table}[t]
\caption{Scenario-balanced agreement contrasts. Estimates are percentage
points; intervals use 10,000 scenario-block bootstraps. Holm $p$ values apply
only to the four prespecified exact tests. All rows use 430 valid runs, 279
agreements, and 12 scenarios.}
\label{tab:headline}
\centering
\small

\begin{tabular}{lrrr}
\toprule
Contrast & Estimate & 95\% CI & Adjusted $p$ \\
\midrule
Korean $-$ English & $+24.8$ & $[15.7,34.7]$ & $.0039$ \\
Strong urgency $-$ none & $+29.9$ & $[20.1,39.6]$ & n/a \\
Language $\times$ urgency & $+16.2$ & $[0.9,32.9]$ & $.1133$ \\
Language $\times$ (cooperative $-$ neutral) & $-48.6$ & $[-66.7,-30.6]$ & $.0039$ \\
Language $\times$ (headstrong $-$ neutral) & $-22.9$ & $[-36.8,-8.3]$ & $.0391$ \\
\bottomrule
\end{tabular}
\end{table}

\subsection{Negotiation dynamics and agreed prices}

Buyers' opening anchors were virtually identical across languages, with normalized first offers differing by only $+0.024$ in Korean (95\% CI $[-0.011,0.056]$).  Later behavior diverged: Korean reduced buyer net
concession by $0.086$ target spans (CI $[-0.137,-0.041]$), increased seller
net concession by $0.349$ spans (CI $[0.237,0.462]$), and reduced the
terminal buyer--seller gap by $0.286$ spans (CI $[-0.400,-0.175]$). Thus the
agreement result is more consistent with different concession dynamics and
gap closure than with a different buyer opening anchor. Mean length differed
by only $+0.41$ turns (CI $[-0.17,1.07]$), while strong urgency shortened
negotiations by $0.94$ turns (CI $[-1.58,-0.33]$). Full offer trajectories are in Appendix~\ref{app:trajectories}. Among 279 agreements, normalized price averaged 0.513 in English and 0.263
in Korean, a $-0.250$ contrast (CI $[-0.315,-0.178]$), favoring buyers. This
gap does not mean Korean negotiations produce better prices for buyers in
general: language changes which negotiations reach agreement, and only two
scenarios supplied agreements in every full factorial cell. We therefore use
price only to describe the selected agreement set.

\subsection{Directional check with a second negotiator model}
We repeated the full design with Qwen3.7 Flash as both agents
(Appendix~\ref{app:qwen-validation}). Agreement was again higher in Korean
($+13.9$ vs.\ Gemini's $+24.8$ points), but its interval reaches down to
zero ($[0.0,28.7]$) and 12 Korean-condition runs were conducted in English.
Sellers again conceded more in Korean ($+0.121$ spans, CI $[0.031,0.222]$)
and the terminal gap was smaller ($-0.147$, CI $[-0.235,-0.065]$). We
therefore read Qwen as directional support, not a replication or a third
confirmatory test.

\subsection{Politeness and pragmatic annotations}
FTA rates were high and similar across languages, unsurprisingly, since
rejecting an offer or countering a price is itself face-threatening. What
differed was mitigation: Korean turns softened the face threat notably more
often than English, most sharply for sellers. Korean dialogue also used a
higher share of positive-politeness strategies for both roles (full rates
and strategy composition by role are in Appendix~\ref{app:pragmatics}).
These labels are compatible with a language-conditioned realization
difference, but neither their observational association with agreement nor
a judge-model label establishes a mechanism.

\section{Discussion}
\label{sec:discussion}

The 24.8-point Korean gain is not uniform across buyer types. Neutral and headstrong buyers drive it, while cooperative buyers were already near ceiling. This divergence appears only after the opening offer, as sellers concede further and the two sides end up closer together in Korean. This matches the failure
mode cross-lingual social-policy fidelity predicts: the persona's wording
stays the same across languages, but its economic effect does not.

The language effect is concentrated in abandonment rather than the turn cap:
English had 39.1\% abandonment versus 15.7\% in Korean. The English neutral
baseline without urgency is unusually low (11.1\% agreement, 88.9\%
abandonment; Table~\ref{tab:cells}). In 28 of its 32 abandonments the buyer
walked away, typically within eight turns, after treating its private target
as a hard limit; Korean neutral buyers did so in 13 of 17. This low baseline
enlarges the no-urgency language gap, so the size of the headline contrast
should be read as specific to this prompt and model. Strong urgency independently increased agreement, but its interaction with
language remained only suggestive (Table~\ref{tab:headline}). All inferential tests treat the 12 scenarios, not the 430 runs, as
independent units. The second-model check preserved the direction of the
agreement, seller-concession, and gap-closure effects, with smaller
magnitudes (Appendix~\ref{app:qwen-validation}). This evaluation is useful because it makes a deployable failure
visible before an agent represents real users: teams
can check whether persona and hardship policies stay behaviorally stable
across the languages they support. The same instability could give users different deals or prices depending
on their interface language; higher agreement is not automatically fairer or
better for users. The experiment must not be used to
stereotype human negotiators; it measures model behavior under translated
prompts. Any deployment audit should report agreement alongside selection into
observed prices and role-specific concessions, not agreement alone.

\section{Limitations and future work}
\label{sec:limitations}
This pilot covers 12 translated U.S. marketplace scenarios, three
repetitions, and one primary model, so it says nothing about Korean markets or
human negotiators. Because the persona and urgency instructions were
translated along with the conversation, part of the language effect could
come from how strongly a translated instruction reads (e.g., the headstrong
prompt in Korean) rather than from the interaction language itself; and
because buyer and seller always share a language, we cannot tell whether the
shift comes from the Korean seller, the Korean buyer, or both. Qwen3.7 Flash
directionally supports the outcome and offer-process findings but was not
scored for pragmatic annotations (Appendix~\ref{app:qwen-validation}).
Conditional prices remain selected on agreement. The politeness labels come
from an LLM judge with only informal spot checks, not formal human
validation, so we treat them as descriptive. Future work should cross
instruction language with interaction language, pair a Korean buyer with an
English seller (and vice versa), and test additional models, languages,
localized markets, and human-calibrated pragmatics. Reproducibility and
exclusion details are in Appendix~\ref{app:reproducibility}.
\clearpage

\bibliographystyle{plainnat}
{\small
\bibliography{references}
}

\clearpage
\appendix

\section{Protocol details}
\label{app:protocol}

\paragraph{Scenarios.} The 12 scenarios were drawn with seed 42, two each from
housing, furniture, cars, bikes, phones, and electronics.

\paragraph{Prompts.} The role, item, target, persona, and urgency text is in
the interaction language. The JSON response-format instruction is in English
in both conditions, and no prompt names the language to respond in
(Appendix~\ref{app:translation}).

\paragraph{Outcomes and exclusions.} An accept ends a run as
\texttt{agreed}, a terminal reject as \texttt{abandoned}, and reaching 16
turns as \texttt{max\_turns\_reached}. A response that failed semantic
parsing was regenerated up to three times. Two runs, both in the
English/strong/headstrong cell, still failed and were dropped without
replacement (Appendix~\ref{app:reproducibility}). The remaining 430 runs
contain 4,580 turns and end in 279 agreements, 118 abandonments, and 33
turn-cap outcomes.

\paragraph{Measures.} We report agreement rate and agreed price for
comparability with prior LLM benchmarking on CraigslistBargain
\citep{agreemate2024}. Normalized prices are not clipped. We also read the
pragmatic labels through Ting-Toomey's Face-Negotiation Theory
\citep{tingtoomey1985conflict}, without testing mediation. Scenario sampling
and the bootstrap both use seed 42.

\FloatBarrier

\section{Agreement outcomes by scenario}
\label{app:scenario-agreement}

Table~\ref{tab:scenario-agreement} reports agreement separately for each
scenario and interaction language. Rates average the factorial conditions and
repetitions available within a scenario.

\begin{table}[ht]
\caption{Agreement by scenario and interaction language. $A/N$ denotes
agreements per valid negotiation. The difference is the Korean-language rate
minus the English-language rate in percentage points. The summary row reports
total counts and scenario-balanced rates.}
\label{tab:scenario-agreement}
\centering
\small
\begin{tabular}{llrrrrr}
\toprule
& & \multicolumn{2}{c}{English} & \multicolumn{2}{c}{Korean} & \\
\cmidrule(lr){3-4}\cmidrule(lr){5-6}
Scenario & Category & $A/N$ & Rate (\%) & $A/N$ & Rate (\%) & Difference \\
\midrule
00 & Furniture   & 9/17  & 52.9  & 12/18 & 66.7  & 13.7 \\
01 & Car         & 8/18  & 44.4  & 16/18 & 88.9  & 44.4 \\
02 & Bike        & 5/18  & 27.8  & 10/18 & 55.6  & 27.8 \\
03 & Phone       & 6/18  & 33.3  & 17/18 & 94.4  & 61.1 \\
04 & Bike        & 7/18  & 38.9  & 9/18  & 50.0  & 11.1 \\
05 & Electronics & 10/18 & 55.6  & 15/18 & 83.3  & 27.8 \\
06 & Car         & 7/17  & 41.2  & 12/18 & 66.7  & 25.5 \\
07 & Housing     & 12/18 & 66.7  & 14/18 & 77.8  & 11.1 \\
08 & Phone       & 13/18 & 72.2  & 17/18 & 94.4  & 22.2 \\
09 & Electronics & 18/18 & 100.0 & 18/18 & 100.0 & 0.0 \\
10 & Furniture   & 7/18  & 38.9  & 15/18 & 83.3  & 44.4 \\
11 & Housing     & 10/18 & 55.6  & 12/18 & 66.7  & 11.1 \\
\midrule
Summary & & 112/214 & 52.5 & 167/216 & 77.3 & 24.8 \\
\bottomrule
\end{tabular}
\end{table}

\FloatBarrier

\section{Full outcome cells}
\label{app:cells}

Table~\ref{tab:cells} keeps abandonment and the 16-turn cap separate. Price is
normalized and conditional on agreement. The two technical exclusions are in
English/strong/headstrong.

\begin{table}[ht]
\caption{Full factorial outcomes. $A/V$ is attempted/valid; $D$, $B$, and $C$
are deal, abandonment, and cap rates. $z(p)$ is conditional on the listed
number of agreements. Every row contains 12 scenarios.}
\label{tab:cells}
\centering
\small
\begin{tabular}{lllrrrrr}
\toprule
Lang. & Urgency & Persona & $A/V$ & $D$ & $B$ & $C$ & $z(p)$ ($n$) \\
\midrule
EN & none & cooperative & 36/36 & 1.000 & 0.000 & 0.000 & .939 (36) \\
EN & none & neutral     & 36/36 & .111 & .889 & 0.000 & .270 (4) \\
EN & none & headstrong  & 36/36 & .139 & .583 & .278 & .705 (5) \\
EN & strong & cooperative & 36/36 & .972 & .028 & 0.000 & .751 (35) \\
EN & strong & neutral     & 36/36 & .389 & .611 & 0.000 & .092 (14) \\
EN & strong & headstrong  & 36/34 & .542 & .236 & .222 & .320 (18) \\
KO & none & cooperative & 36/36 & .972 & 0.000 & .028 & .704 (35) \\
KO & none & neutral     & 36/36 & .500 & .472 & .028 & .251 (18) \\
KO & none & headstrong  & 36/36 & .278 & .444 & .278 & .122 (10) \\
KO & strong & cooperative & 36/36 & 1.000 & 0.000 & 0.000 & .428 (36) \\
KO & strong & neutral     & 36/36 & .972 & .028 & 0.000 & .040 (35) \\
KO & strong & headstrong  & 36/36 & .917 & 0.000 & .083 & .034 (33) \\
\bottomrule
\end{tabular}
\end{table}

\FloatBarrier

\section{Offer trajectories}
\label{app:trajectories}
\begin{figure}[ht]
  \centering
  \includegraphics[width=.9\linewidth]{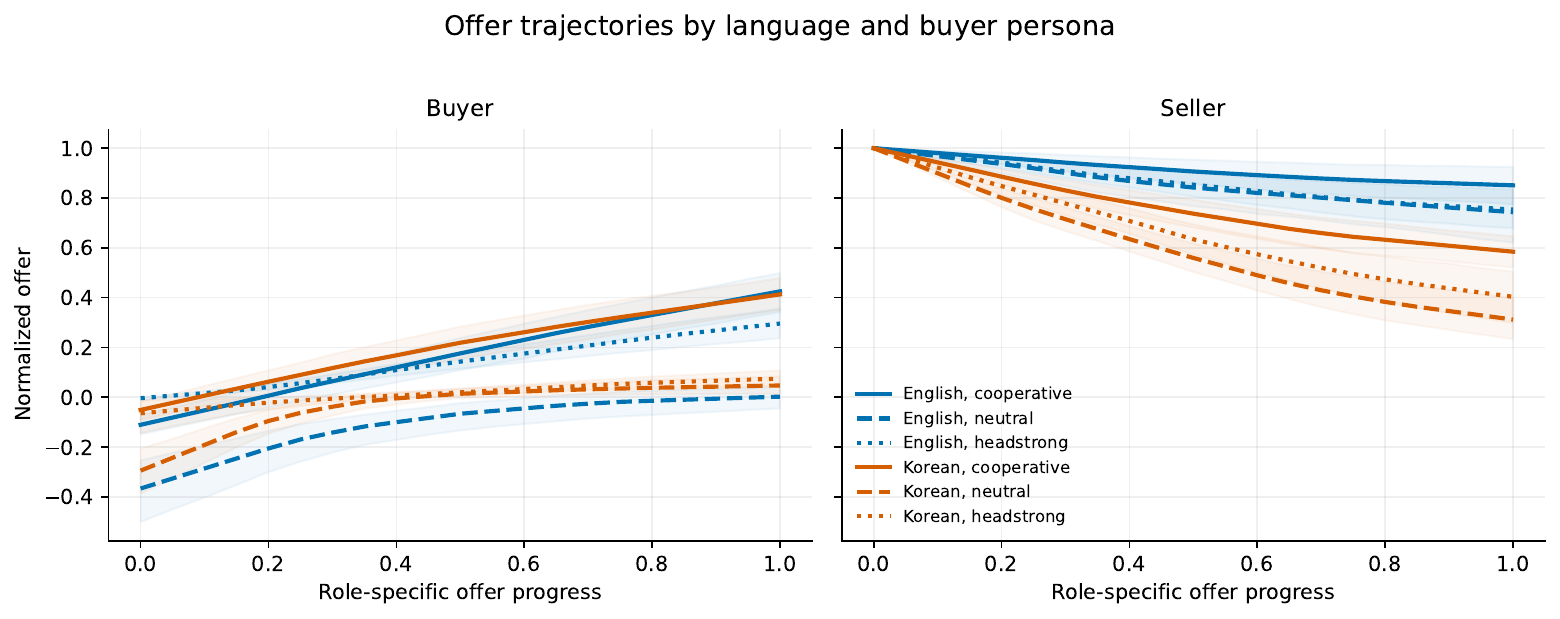}
  \caption{Normalized buyer and seller offers over negotiation progress,
  aggregated with equal scenario weight. Trajectories are descriptive because
  negotiations end at different times and for different reasons.}
  \label{fig:trajectories}
\end{figure}

\FloatBarrier

\section{Second-model directional check}
\label{app:qwen-validation}

We repeated the complete $2\times2\times3$ design with Qwen3.7 Flash
(\texttt{qwen/qwen3.7-flash}) as both buyer and seller, routed through
OpenRouter at temperature 0.2. This is the same model that served as the
pragmatic judge for the Gemini transcripts; it did not judge its own
negotiations.
The only change from the primary experiment was the negotiator model. The 12 scenarios, English and Korean materials, social policy prompts, private target values, three repetitions, role order, and 16-turn stopping rule all stayed the same.
All 432 planned Qwen negotiations were
present and parsable: 180 ended in agreement, 237 in abandonment, and 15 at
the turn limit. 
Because this Qwen set has no pragmatic judge labels, the
comparison is limited to terminal outcomes and structured offers.

Agreement was 34.7\% in the English condition (75/216) and 48.6\% in the
Korean condition (105/216), a scenario-balanced increase of 13.9 percentage
points (Table~\ref{tab:qwen-validation}). Strong seller urgency also retained
a large association with agreement, increasing it by 44.4 points. The Korean
condition had higher agreement within every buyer persona: 63.9\% versus
48.6\% for cooperative buyers, 36.1\% versus 22.2\% for neutral buyers, and
45.8\% versus 33.3\% for headstrong buyers. Qwen distributed the language
contrast more evenly across personas than Gemini, indicating that the exact
size of persona-specific contrasts varies across model families.

\begin{table}[ht]
\caption{Scenario-balanced estimates from the Qwen directional check. Language
contrasts are Korean minus English; the urgency contrast is strong minus none.
Agreement and urgency use all 432 runs, while offer-process rows use the 425
runs with analyzable price-bearing sequences. Every row covers 12 scenarios.}
\label{tab:qwen-validation}
\centering
\small
\begin{tabular}{lrr}
\toprule
Contrast & Estimate & 95\% CI \\
\midrule
Agreement (percentage points) & $+13.9$ & $[0.0,28.7]$ \\
Strong seller urgency (percentage points) & $+44.4$ & $[37.5,51.4]$ \\
First buyer offer & $+0.017$ & $[-0.004,0.038]$ \\
Buyer net concession & $+0.009$ & $[-0.018,0.033]$ \\
Seller net concession & $+0.121$ & $[0.031,0.222]$ \\
Terminal buyer-seller gap & $-0.147$ & $[-0.235,-0.065]$ \\
\bottomrule
\end{tabular}
\end{table}

\begin{figure}[ht]
  \centering
  \includegraphics[width=\linewidth]{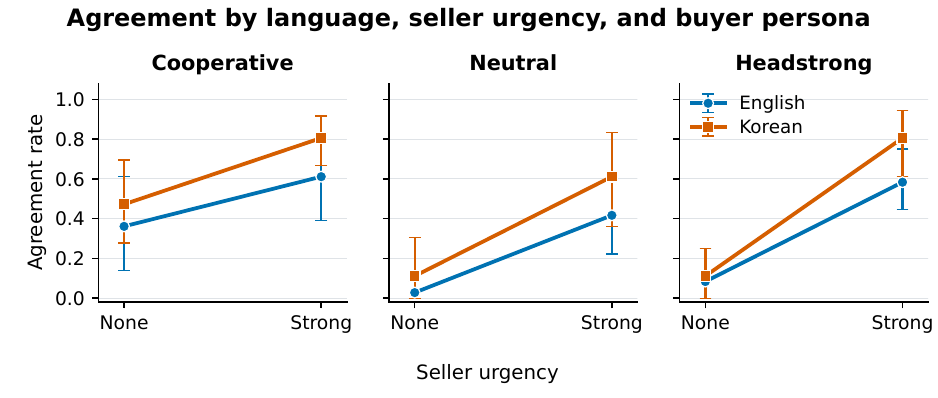}
  \caption{Scenario-balanced Qwen agreement by language, seller urgency, and
  buyer persona across all 432 assigned-condition runs. Error bars show 95\%
  scenario-block bootstrap intervals.}
  \label{fig:qwen-agreement}
\end{figure}

\begin{figure}[ht]
  \centering
  \includegraphics[width=.9\linewidth]{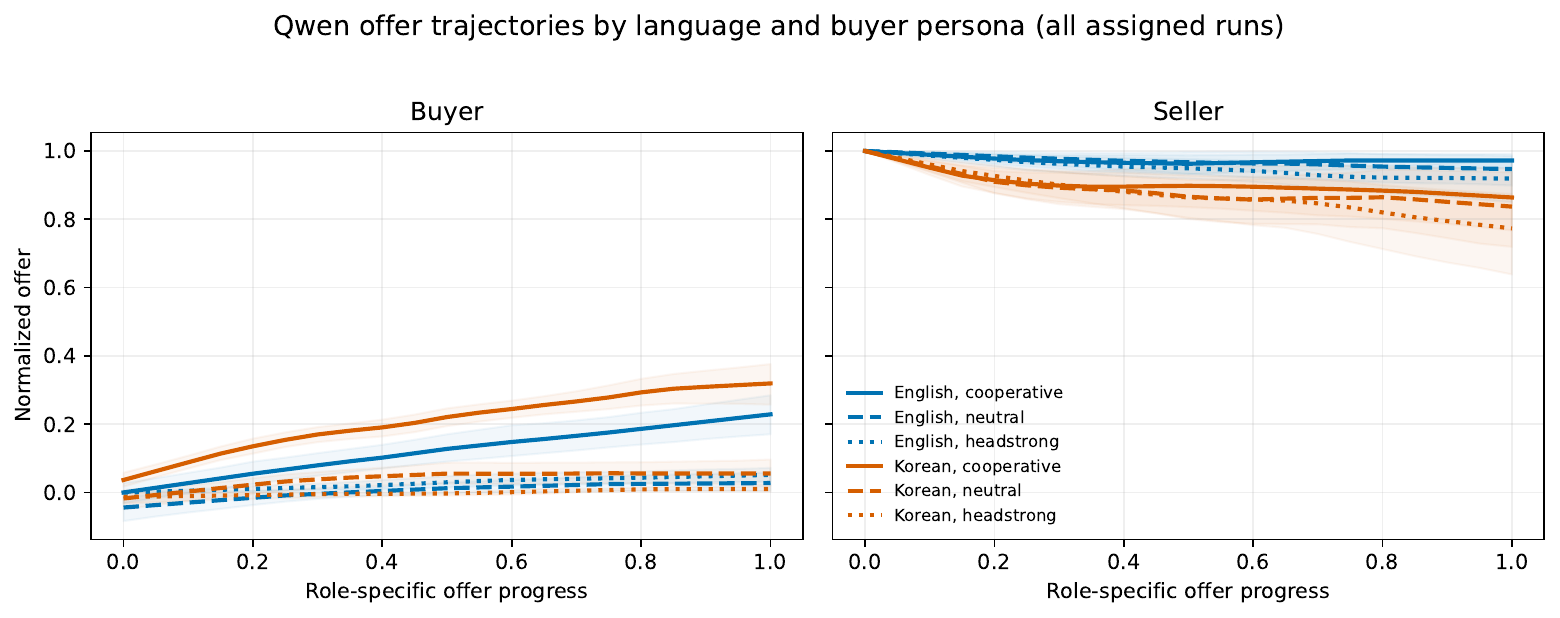}
  \caption{Normalized Qwen buyer and seller offers over negotiation progress,
  aggregated with equal scenario weight. The figure includes 425 runs with
  analyzable offer sequences; seven one-turn acceptances contained no offer
  trajectory.}
  \label{fig:qwen-trajectories}
\end{figure}

The offer data show the same process signature as the primary experiment.
Buyer opening anchors remained similar across languages, while sellers in the
Korean condition made larger net concessions and negotiations ended with
smaller buyer-seller gaps. Figure~\ref{fig:qwen-trajectories} shows that this
pattern develops over the course of the negotiation rather than at the opening
anchor.

Twelve Korean-condition runs produced English-only dialogue. Excluding these
runs left the agreement contrast unchanged at $+13.9$ points, and a
conservative schema-compliant sensitivity estimate was $+14.4$ points. The
direction therefore holds across the full analysis and
both sensitivity checks, although the lower bound of the agreement interval
is at zero. We limit the cross-model comparison to agreement and
offer-process outcomes because no Qwen scenario supplied agreements in every
factorial cell needed for an equally weighted conditional price contrast.
Descriptively, the condition-balanced mean agreed price (excluding the 12
English-output runs) was 0.580 target spans in English and 0.392 in Korean
($-0.188$, vs.\ Gemini's $-0.250$), but this comparison carries no interval
and is selected on agreement.

\FloatBarrier

\section{Pragmatic summaries}
\label{app:pragmatics}

\begin{figure}[ht]
  \centering
  \includegraphics[width=.78\linewidth]{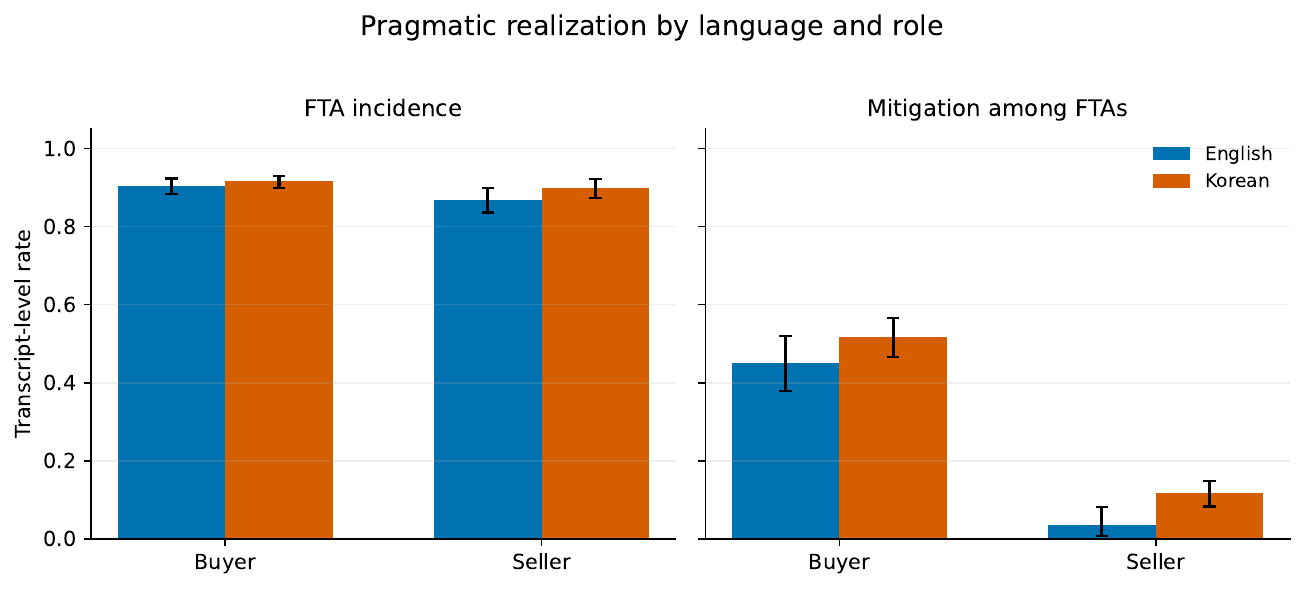}
  \caption{Transcript-level FTA and mitigation rates by language and role.
  Bars are 95\% scenario-block bootstrap intervals. These judge-derived
  measures are descriptive and are not used for causal mediation.}
  \label{fig:pragmatics}
\end{figure}

\begin{table}[ht]
\caption{Judge-derived pragmatic rates. Counts are valid transcripts; all rows
cover 12 scenarios. Mitigation is conditional on judge-labeled FTAs.}
\label{tab:pragmatics}
\centering
\small
\begin{tabular}{llrrrr}
\toprule
Language & Role & Runs & Agreements & FTA rate & Mitigation rate \\
\midrule
English & buyer  & 214 & 112 & .904 & .449 \\
Korean  & buyer  & 216 & 167 & .916 & .516 \\
English & seller & 214 & 112 & .868 & .036 \\
Korean  & seller & 216 & 167 & .899 & .117 \\
\bottomrule
\end{tabular}
\end{table}

\begin{figure}[ht]
  \centering
  \includegraphics[width=.82\linewidth]{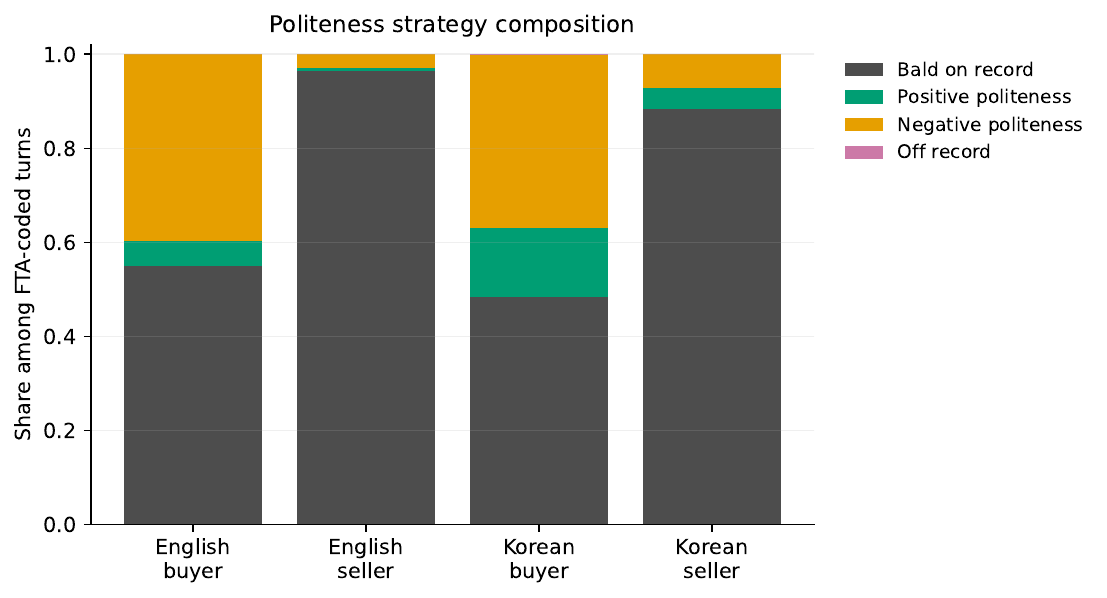}
  \caption{Composition of judge-labeled Brown and Levinson politeness
  strategies by interaction language and role. Shares are conditional on an
  FTA and are descriptive.}
  \label{fig:politeness-composition}
\end{figure}

\begin{table}[ht]
\caption{Scenario-balanced mean politeness-strategy shares among judge-labeled
FTAs.}
\label{tab:politeness-composition}
\centering
\small
\begin{tabular}{llrrrr}
\toprule
Language & Role & Bald & Positive & Negative & Off record \\
\midrule
English & buyer  & .551 & .053 & .396 & .000 \\
Korean  & buyer  & .484 & .146 & .369 & .002 \\
English & seller & .964 & .007 & .030 & .000 \\
Korean  & seller & .883 & .045 & .071 & .000 \\
\bottomrule
\end{tabular}
\end{table}

\FloatBarrier

\section{Model and API configuration}
\label{app:model-config}

Table~\ref{tab:model-config} reports the generation settings recoverable from
the collection code and stored outputs. Credentials and cloud-project
identifiers are omitted. The judge model slug is stored in every judged file;
the raw negotiation files do not serialize the negotiator model or API
settings.

\begin{table}[ht]
\caption{Recorded model and API configuration. Unspecified settings were not
passed by the code and therefore used provider defaults.}
\label{tab:model-config}
\centering
\small
\setlength{\tabcolsep}{4pt}
\begin{tabular}{p{.18\linewidth}p{.36\linewidth}p{.36\linewidth}}
\toprule
Setting & Negotiating agents & Pragmatic judge \\
\midrule
Model & \texttt{gemini-3.6-flash} for both buyer and seller &
\texttt{qwen/qwen3.7-flash} \\
API route & Vertex AI through \texttt{google-genai} with Application Default
Credentials; the code defaults to the \texttt{global} location & OpenRouter's
OpenAI-compatible endpoint \\
Role setup & Separate buyer and seller system prompts using the same model
slug & One discourse-analyst instruction containing the complete transcript \\
Temperature & 0.2 & 0.2 \\
Input construction & All preceding turns, mapped to \texttt{model} or
\texttt{user} relative to the speaking role, followed by a user request to
continue & No prior chat history; the full transcript and coding taxonomy were
placed in one system message \\
Output contract & Text instructed to contain one JSON object with utterance,
action, and price & Text instructed to contain one JSON array with turn, role,
FTA, and strategy labels \\
Call unit & One generation per turn, seller first, alternating roles, with a
16-turn cap & One generation per valid transcript, labeling every turn at once \\
Model revision & Provider alias only; no dated revision was pinned & Provider
alias only; OpenRouter backend route and revision were not recorded \\
Other decoding & \multicolumn{2}{p{.72\linewidth}}{No generation seed,
top-$p$, top-$k$, maximum output length, stop sequence, candidate count,
response schema, or safety override was supplied.} \\
\bottomrule
\end{tabular}
\end{table}

For negotiation turns, semantic JSON failures triggered up to three fresh
generation attempts. Within each such generation, the Vertex wrapper and
OpenRouter client each retried transient errors (rate limits, timeouts,
connection errors, and similar) up to six times with increasing backoff,
honoring any retry delay the API returned; SDK-internal retry behavior was
not overridden. A malformed judge JSON array failed that job rather than
retrying automatically; the resume-safe command could later rerun failed
files.

The original transcript schema did not record collection timestamps, request
identifiers, resolved model revisions, runtime package versions, effective
Vertex project or location, OpenRouter backend route, worker count, prompt or
configuration hashes, per-call retry counts, or the generating code commit.
These values cannot be backfilled reliably. Seed 42 selected scenarios and
controlled the analysis bootstrap; it was not supplied as a model-generation
seed.

\FloatBarrier

\section{Compute, artifacts, and data provenance}
\label{app:resources}

\paragraph{Experiments compute.}
All model inference ran on hosted providers (Vertex AI for Gemini, OpenRouter
for Qwen3.7 Flash as negotiator and judge); the local host used only CPU, with no GPU or accelerator, to
orchestrate concurrent HTTP requests, defaulting to eight concurrent worker
threads for negotiation generation and judging alike. The reported dataset
comprises 4,580 analyzed Gemini negotiation turns (432 attempted
negotiations), 3,266 analyzed Qwen negotiation turns (432 attempted
negotiations), and 430 judge requests (one call per transcript). Wall-clock
span from the first to the last written file was 7.0 hours (Gemini
negotiation), 8.0 hours (Qwen negotiation), and 83.5 hours (judging); these
totals are inflated by unattended pauses between working sessions, which we
identify as any gap between consecutive completions exceeding 15 minutes,
nearly 60$\times$ the typical gap during active processing (median
$\le$7s, 90th percentile $\le$23s across all three workloads). Excluding
these pauses, dense processing time was approximately 1.2 hours (Gemini),
1.1 hours (Qwen), and 0.8 hours (judging), totalling approximately 3.1
hours of active compute across roughly 8,300 hosted API calls. An
additional 91 negotiation attempts failed transiently during the Gemini run
(connection resets coinciding with host sleep) and were retried under the
same resume-safe procedure; these are not double-counted in the totals
above. Two earlier pilot batches were superseded and excluded from the
reported results: a 36-negotiation smoke test under an earlier 10-turn cap,
and a 12-negotiation smoke test using OpenRouter as the negotiator, both
prior to settling on the Vertex AI/16-turn configuration described in
Section~\ref{sec:method}. Provider-side hardware, transport retries, and
token totals were not exposed or retained; the wall-clock figures above are
reconstructed from file write times.

\paragraph{Analysis compute.}
The executed notebook used Python 3.12.3 on a CPU-only Windows host with an
Intel Core i5-9300H, eight logical processors, and 16~GB memory; a clean run
took 25.4 seconds. The 2.2~MB inputs produced 44 artifacts occupying 4.7~MB
with NumPy 2.2.6, pandas 2.3.3, and Matplotlib 3.10.6. The notebook and README
provide checks, hashes, dependencies, and exact commands; two clean runs
matched on all 26 tabular and audit outputs. Code and the analysis notebook
are available at \textcolor{red}{[https://github.com/saeheeeom/polite-bargain]}; transcripts and
annotations at \textcolor{red}{[https://huggingface.co/datasets/polite-bargain/polite-bargain]}.

\paragraph{Source data and terms.}
CraigslistBargain is credited to its original authors
\citep{he2018decoupling}. We used the
\texttt{stanfordnlp/craigslist\_bargains} Hugging Face distribution, accessed
August 18, 2026 \citep{stanfordnlp2026craigslist}. Its current dataset record
does not specify a license. Model calls remain subject to the applicable Google
Cloud, Vertex AI, and OpenRouter service terms.

\FloatBarrier

\section{Exclusions and reproducibility}
\label{app:reproducibility}

\begingroup\raggedright
The parser excluded exactly two files after three failed semantic parse
attempts at a turn: \path{scenario-00_en-strong-headstrong_rep1.json}
(five preceding valid turns) and
\path{scenario-06_en-strong-headstrong_rep0.json} (14 preceding valid turns).
The corresponding judge files do not exist by design; every one of the
430 valid transcripts has exactly one aligned judge file.\par
\endgroup

The notebook validates the design grid, schemas, roles, actions, prices,
outcomes, and judge alignment before exporting results. Two clean executions
with seed 42 matched on all 26 tabular and audit artifacts. Release locations
for code, transcripts, and annotations are listed in
Appendix~\ref{app:resources}.

\FloatBarrier

\section{Translation and prompt materials}
\label{app:translation}

Scenario titles and descriptions, and the persona and urgency instructions,
were machine translated with \texttt{gemini-3.6-flash} and checked in one
line-by-line pass against the English source by a single fluent Korean
speaker on the author team; there was no second annotator or back-translation
study. The pass made two changes. First, in 4 of 12 descriptions the
translation model had appended marketplace-style closing lines with no English
source (e.g., an invitation to message the seller); these were removed to
restore one-to-one content, since they would have made only the Korean
listings read as warmer. Second, all descriptions initially used the same
sentence-final polite register; 4 randomly chosen descriptions (seed 42)
were rewritten in the noun-ending style common in Korean listings to avoid a
uniformly chatty register. U.S. locations and dollar prices were kept.

Table~\ref{tab:prompts} gives the exact persona and urgency strings. Neutral
persona and no urgency add no text in either language. Each role prompt
opens with a one-sentence statement of role, item, listing price, and private
target in the interaction language (Korean: ``\begin{CJK}{UTF8}{mj}당신은 다음 물건을 구매하기 위해
협상 중입니다: \ldots{} 정가: \ldots{} 목표 가격: \ldots\end{CJK}''), followed by the
persona or urgency string and an English JSON response-format instruction
shared by both languages.

\begin{table}[ht]
\caption{Persona (buyer) and urgency (seller) instructions as used in each
language.}
\label{tab:prompts}
\centering
\small
\begin{CJK}{UTF8}{mj}
\begin{tabular}{p{.15\linewidth}p{.38\linewidth}p{.38\linewidth}}
\toprule
Condition & English & Korean \\
\midrule
Cooperative & You prioritize reaching a deal that works for both sides
quickly. You are willing to make concessions readily rather than prolong the
negotiation. & 당신은 양쪽 모두에게 괜찮은 합의를 빠르게 이루는 것을 중요하게
생각합니다. 협상을 오래 끌기보다는 기꺼이 양보하는 편입니다. \\
\addlinespace
Headstrong & You anchor firmly on your target price. You concede slowly and
reluctantly, and only when it seems necessary to avoid losing the deal
entirely. & 당신은 목표 가격을 확고하게 고수합니다. 거래가 완전히 무산될 것 같을
때만, 그것도 천천히 마지못해 양보합니다. \\
\addlinespace
Strong urgency & You need the cash quickly. & 당신은 급하게 현금이 필요합니다. \\
\bottomrule
\end{tabular}
\end{CJK}
\end{table}

\clearpage
\end{document}